\documentclass[conference]{csce}
\usepackage{comment}
\usepackage[hmargin=.75in,vmargin=1in]{geometry}
\usepackage[american]{babel}
\usepackage[T1]{fontenc}
\usepackage{times}
\usepackage{algorithm}
\usepackage{algorithmicx}
\usepackage{algpseudocode}	 
\usepackage{mathrsfs}
\usepackage{bm}
\usepackage{color}
\usepackage{slashbox}
\usepackage{setspace}
\usepackage{xspace}
\usepackage{times}
\usepackage{amsmath}
\usepackage{booktabs}
\usepackage{textcomp}
\usepackage{epsfig,graphicx}
\usepackage{xcolor}
\usepackage{amsfonts,amsmath,amssymb}
\usepackage{booktabs}

\title{ \bf %Optimization of Existing Methodologies, Optimal Parameter Tuning for Zero-Shot Learning or (
On Parameter Tuning in Meta-learning for Computer Vision}
\author{Farid Ghareh Mohammadi$^1$,  M.  Hadi  Amini$^2$,  and Hamid R.  Arabnia$^1$\vspace{0.2in}\\  1:  Department of Computer Science,  Franklin College of Arts and Sciences \\ University of Georgia,  Athens,  Georgia,  30602  \\
2:  School of Computing and Information Sciences,  College of Engineering and Computing\\Sustainability, Optimization, and Learning for InterDependent networks laboratory (solid lab)  \\Florida International University,  Miami,  FL 33199 \\
Emails :  farid.ghm@uga.edu,    moamini@fiu.edu,  hra@cs.uga.edu}

\begin{document}

\maketitle                        %%%% To set Title and Author names.

\begin{abstract}%%%% Replace with your abstract.
 Learning to learn plays  a pivotal role in meta-learning (MTL) to obtain an optimal learning model. In this paper, we investigate  image recognition for unseen categories of a given dataset with limited training information. We deploy a zero-shot learning (ZSL) algorithm to achieve this goal. We also explore the effect of parameter tuning on performance of semantic auto-encoder (SAE). We further address   the parameter tuning problem for meta-learning, especially focusing on zero-shot learning. By combining
different embedded parameters, we improved the accuracy of tuned-SAE. Advantages and disadvantages of parameter tuning and its application in image classification are also explored.  
 
\end{abstract}

\vspace{1em}

\section{Introduction}
\textit{Motivation}: Computer vision algorithms are essential to enable modern functionalities in future smart cities \cite{amini2019sustainable} \cite{profarabniaSmartcity}, including face recognition \cite{ProfArabnia5} and automatic license  plate recognition. Image classification stands as the primary challenging problem in computer vision \cite{mohammadi2012survey} \cite{mohammadi2014new} \cite{mohammadi2017region} \cite{mohammadi2014image}. Computer vision provides important applications such as visual and infrared sensor data-based obstacle detection for the visually impaired \cite{jafri2017visual}. Within computer vision, deep learning is an advanced learning tools based on using a training dataset to obtain a high performance on a testing dataset \cite{profarabnia3} \cite{profarabnia4DeepLearning}. Furthermore,  auto-encoders have become recent challenging work in computer vision \cite{wang2019improved}; therefore, we focus on semantic auto-encoder in this paper. The ability to learn new tasks efficiently by leveraging prior experience from related tasks plays a main role in the world of artificial intelligence, especially learning. Meta-Learning (MTL) is the most advanced machine learning algorithm because it acquires the potential ability to learn as efficiently as possible from prior information. MTL first was presented as early as 1987 by Schmidhuber \cite{Schmidhuber1987Learning}, and recent state-of-the-art studies \cite{finn2017model} \cite{ rajeswaran2019meta}  \cite{chen2019modular} illustrate how MTL perfectly learns from a limited training dataset. In this paper, we investigate a task of image recognition to classify given images of unseen categories in the testing dataset for which we had a lack of examples in the training dataset. We choose zero-shot learning (ZSL) for the process of unseen image recognition \cite{lampert2013attribute} to overcome this limitation properly. 

Zero-shot learning is known as attribute-based learning, defined as the process of learning to recognise unseen objects. ZSL emphasizes learning of a new distribution of seen classes,
given meta description of the seen categories, and seeks correlations with existing seen categories in a training dataset. This means that ZSL no longer needs to have any seen samples of unseen classes before evaluating a performance of predicting unseen classes.  
In recent years, ZSL \cite{Ding2019} \cite{ zhang2019deep} \cite{ chen2019modular} \cite{ finn2017model} \cite{ xian2017zero} has been an active, challenging and hot research topic in advanced computer vision, machine learning, and medical data analysis \cite{larochelle2008zero}. Drug discovery \cite{larochelle2008zero}, image classification \cite{zhang2019deep} \cite{finn2017model} and meta-sense \cite{Mohammadi2019sensor}
are examples of such research studies. Furthermore, ZSL has penetrated other domains like human action recognition, networks \cite{zhang2019deep}, etc. In \cite{zhang2019deep}, researchers provided  comprehensive information about ZSL applications in computer and mobile networks.

ZSL is a promising MTL algorithm that behaves like  human cognitive activity. Here, we discuss emerging problems with ZSL. First, semantic space representations, which include attributes ('A') and word vector ('W'), are critical for understanding and learning from the seen categories to apply on unseen categories, but these representations appear challenging to ascertain a high performance \cite{Ding2019}. Human can only understand visual objects and attributes based on image explanations to recognise new images. However, the explanations do not yield distinguishable results \cite{parikh2011interactively}, even traditional machine learning algorithms  will not have promising performances \cite{duan2012discovering}. Second, ZSL provides a mapping model selection to work  with the unseen classes. The most important feature of ZSL consists of learning a compatible visual-semantic or attribute-based function and its ability to semantically represent objects. All in all, the more complex functions we have, the higher the risk of over-fitting, which yields poor results applied  on the unseen classes. Whereas, simple linear function currently yields poor classification accuracy on the seen classes and does not adapt accurately on the unseen classes.

\textit{Contribution}: In this paper, we address the two  aforementioned emerging problems in ZSL and present one promising solution. The main contribution of this paper is to  provide a linear optimal function using tuned parameters to reach the most promising  classification result that outperforms the most state-of-the-art work. We illustrate a detailed procedure of the work for meta-Mapping which is inspired from semantic auto-encoder (SAE) in algorithm \ref{algorithm: MTL-Meta-Mapping}. In the given meta-mapping, we extend the work presented by Kodirov \emph{et al} \cite{kodirov2017semantic}

 \textit{Algorithm}
 Semantic Auto-Encoder (SAE) \cite{kodirov2017semantic} is an advanced linear mapping auto encoder that aims to find an optimal mapping function ($\mathscr{W}$) to recognise and classify unseen classes. Algorithm \ref{algorithm: MTL-Meta-Mapping} illustrates a comprehensive procedure of SAE. First, in this paper, we develop SAE properly, and in the second step, we optimise the algorithm leveraging the tuning of a few embedded parameters of SAE. Note that in addition to the meta-learning for computer vision application, parameter tuning plays a pivotal role in ensuring  convergence of several algorithms \cite{hadiaminidist,aminiaccess2019}.  
 
\begin{algorithm}[H]
	\caption{Implementation of optimized zero-shot learning (SAE) \cite{kodirov2017semantic}}
	\begin{algorithmic}[1]
	%	\State{\bf{Input:}}A batch set of training input $(\mathscr{(X ,Y)})$  and  $Batch_{size}$ , test set$(\mathscr{(X' ,Y')})$ .
	%	\State{\bf{Output:}} The best meta-mapping Matrix $(\mathscr{W})$ for zero-shot learning
 	\Require A batch set of training input $(\mathscr{X ,Y})$  and  $training_{size}$
 	\Ensure The best mapping matrix $(\mathscr{W})$ for zero-shot learning
 		\State{\bf{Tuning embedded parameters}}
 		 \rlap{\smash{$\left.\begin{array}{@{}r@{}r@{}r@{}}\\{}\end{array}\color{red}\right\}%
          \color{red}\begin{tabular}{l}We contribute here\end{tabular}$}}
		\State{\bf{Begin Training}}

	        \For{ t=0 $\cdots training_{size}$}
	            \State {Learn $\mathscr{W}$:$\mathcal{Y}$ $\Leftarrow$ $\mathscr{W}$ $\mathscr{X}$}
	           
                \State{$Err_{dst.}$=$\mid\mid $ $\mathscr{X}$ - $\mathscr{W}$$\mathscr{W}'$  $\mathscr{X}$ $\mid\mid _F$}
                 \rlap{\smash{$\left.\begin{array}{@{}r@{}r@{}r@{}}\\{}\\{}\\{}\\{}\end{array}\color{red}\right\}%
          \color{red}\begin{tabular}{l}We learn  $\mathscr{W}$ \\ until $Err_{Dst.} \leadsto 0$\end{tabular}$}}
	            \State {Optimize  $(\mathscr{W})$ :$\mathscr{W}$ =$\frac{\mathscr{C}}{\mathscr{A}+\mathscr{B}}$  }
	            \State {\bf{Return}  $\mathscr{W}$ and$ \mathscr{ W^T}$ }

	         \EndFor
	      \State{\bf{End Training}}		 
	       \State{\bf{Begin Testing }}  
	         \State {Compute acc. on Unseen Classes }
	         \For{ t=0 $\cdots test_{size}$} 
	          
    	        \State{ $\Delta = \mid \mid Pred - Ground_{T}\mid \mid$}
    	       	 \rlap{\smash{$\left.\begin{array}{@{}r@{}r@{}r@{}}\\{}\\{}\\{}\\{}\\{}\\{}\end{array}\color{red}\right\}%
          \color{red}\begin{tabular}{l}We test \\unseen classes\\ maximize the \\performance of SAE \end{tabular}$}}
    		     \State{Minimise  $\Delta$ }
    		     \If{($\Delta$ is minimum)}
    		      \State {$\text{Performance}+=\frac{1}{test_{size}}$ } 
    	        \EndIf  
    	        
    	        \EndFor
	       \State{\bf{End Testing}}		
	        		
	\end{algorithmic}
	\label{algorithm: MTL-Meta-Mapping}
\end{algorithm} 

\textit{Organization}: The rest of this paper is organized as follows. First we review related works regarding the use of ZSL for recognition of unseen classes. In Section 3, we state 
preliminary know-ledge of meta-learning before expressing a suggested meta-mapping (ZSL) algorithm. We then present experimental results and discussion, followed by conclusions.

\section{Related Works}
%10-15 CV papers  including the papers that our main reference compared their work with

In this section, we study related works on the use of zero-shot learning and presented previous examination on the evaluating unseen classes. While the number of new zero-shot learning methods increasing yearly, the criteria to examine all methods are not the same \cite{xian2017zero}. This makes our evaluation the methods particularly difficult.

We present meta-learning to overcome the traditional machine learning limitation, which is when the learned model fails to predict testing data when the class is not already trained in training phases. Meta-learning, specially zero-shot learning overcomes this limitation by recognizing instances of unseen classes, which may not have been trained during training. In this section, we discuss related ZSL research studies. 
\begin{table*}[t]
\center
  \caption{Comparing  related datasets ( SS-D refers to dimension of semantic space dimension)}
  \vspace{0.5cm}
  \renewcommand{\arraystretch} {1}
  \hspace{-1cm}\begin{tabular}{|c|c|c|c|c|}
    \hline 
    \backslashbox{\textbf{Features}}{\textbf{Datasets}}  & AwA \cite{lampert2013attribute}%& SUN
    & CUB \cite{wah2011caltech} &ImNet-2 \cite{fu2016semi}\\ 
         \hline
      images & 30,475&% 14,340
      11,788&218,000 \\
     total classes & 50&%717
     200&1360\\
     seen classes & 40&%645
     150&1000\\
     %validation classes & 13&65&50&\\
     unseen classes & 10&%72
     50&360\\
      SS-D & 85&%102
      312& 1000\\
                    \hline
    \end{tabular}
       \label{tab:dataset}
  \end{table*}
Lampert \emph{et al} \cite{lampert2013attribute} proposed an attribute-based prediction method for ZSL towards object recognition. They introduced a direct attribute prediction (DAP) method to do classification that learns from  seen classes (y) and is tested on unseen classes ($y'$). The authors stated that leveraging attributes enables us to have highly accurate and cost effective knowledge transfer between seen classes in a training dataset and unseen classes in a testing dataset. 
Romera and Torr \emph{et al} \cite{romera2015embarrassingly} presented embarrassingly simple zero-shot learning (ESZSL) and evaluated this with DAP, which is a baseline algorithm. Then, they updated compatibility of the learning algorithm by adding a regularization term.

Zhang and Saligrama \cite{zhang2015zero} \cite{zhang2016zero} introduced a new ZSL based on semantic similarity embedding (SSE) and joint latent similarity embedding (JLSE) respectively. These researchers \cite{zhang2016zero} formulated  zero-shot recognition (ZSR) as a binary classification problem. They studied a framework leveraging dictionary learning to entirely learn the parameters of the model, that finally led to a class-free classifier.

Akata   \emph{et al} \cite{ akata2015evaluation} presented an image classification technique called, structured joint embedding (SJE). Meanwhile, Changpinyo \emph{et al} \cite{changpinyo2016synthesized} introduced a synthesized classifier (SYNC) method using combined semantic descriptions of $(A+W)$ to provide a higher performance on image recognition in comparison with DAP.
\begin{comment}

\begin{table}[t] 
\center
  \caption{Comparing the related datasets}
  \vspace{0.5cm}
  \renewcommand{\arraystretch} {1}
  \hspace{-1cm}\begin{tabular}{|c|c|}
    \hline 
     \textbf{Parameter}  &Value  \\ 
         \hline
      Lamda $(\mathscr{W})$ &  4 \\
     HITK & 2 $\cdots$ 10\\
     dist & no complimentary using pdist2\\
     dist-Kernel &  minkowski\\
     %validation classes & 13&65&50&\\
     Sorting type & ascend\\
      
                    \hline
    \end{tabular}
     \label{table:Tunedparameter}
  \end{table}
\end{comment}
From there, Bucher \emph{et al} \cite{bucher2016improving} proposed a method, which leveraged adding visual features into attribute space, and then learned a
metric to minimize the inconsistency by maximizing adaptability of the semantic embedding. 
Shigeto \emph{et al} \cite{shigeto2015ridge} found that a least square regularised mapping function does not yield a good result for the hubness problem. Thus, they proposed regression and CCA-based approaches to ZSL to compute reverse regression, which means embedding class prototypes into the visual feature space. SAE \cite{kodirov2017semantic} proposed by Kodirov \emph{et al} a semantic auto-encoder 
to regularize the learned model by mapping the image feature to semantic space. Although Xina \emph{et al} \cite{Xian_2019_CVPR} proposed a feature generating framework for $any-shot$ learning called, f-VAEGAN-D2, there is room for improvement. A lot of work have done for small datasets  \cite{lampert2013attribute} \cite{zhang2015zero} \cite{zhang2016zero} \cite{romera2015embarrassingly} \cite{lampert2013attribute} \cite{changpinyo2016synthesized} \cite{ akata2015evaluation} \cite{zhang2015zero} \cite{zhang2016zero}  \cite{bucher2016improving} \cite{shigeto2015ridge} \cite{kodirov2017semantic}, however, just few methods are proposed for large datasets \cite{norouzi2013zero} \cite{fu2016semi}.

\textbf{Hybrid embedding system}: Norouzi \emph{et al} \cite{norouzi2013zero} proposed a hybrid image embedding system, referred to as convex combination of semantic embeddings (ConSE), to deal with \textit{n-way}  image classification that lies in between independent classifier learning and compatibility learning frameworks. ConSE takes images and maps them into the semantic embedding space using convex combination of the class label embedding vectors. Furthermore, 
Fu and Sigal \cite{fu2016semi} presented a learning called semi-supervised vocabulary-informed learning (SS-Voc).
  
\section{Meta-learning for Computer Vision:  Preliminaries and Algorithms}
 \subsection{Preliminaries}
In order to cover all available classes, we need machine learning and evolutionary algorithms try to solve high dimensionality problems of large datasets such as curse of dimensionality (CoD) \cite{mohammadi2014new} \cite{mohammadi2014image}. But, machine learning still  cannot learn all samples when there are few instances per class. Meta-learning alleviates this problem by providing an advanced learning process. Meta-learning has three important promises for computer vision problems, specifically image classification and image recognition: 1) few-shot learning (FSL), 2) one-shot learning (OSL), and 3) zero-shot learning (ZSL). The crux of FSL is to learn a meta-learner to understand categories with boundaries without more than few examples of each category. FSL or \textit{k}-shot learning takes k samples for each category in the training phase. Algorithm \ref{tab: MTL}, which is combination of FSL and OSL, presents semantic pseudocode of model agnostic meta-learning (MAML) which was proposed by Finn \cite{finn2017model}. This algorithm added one more step, which better illustrates how MTL can overcome traditional machine learning limitation using a meta-learner to optimize the learner with gradient decent optimization.

\begin{algorithm}[H]
	\caption{Implementation of Meta-learning (MAML) \cite{finn2017model}}
	\begin{algorithmic}[1]
		\Require A batch set of input targets (I,Y) and  $Batch_{size}$
		\Ensure The best meta-learner $F(\theta')$ for few shot learning and an optimal mapping matrix $(\mathscr{W})$ to zero-shot learning
		\State{\bf{Begin}}
	    \While {work is not done}
	    
	        \For{ t=0 $\cdots Batch_{size}$}
	            \State {Learn from training batch set}
	            \rlap{\smash{$\left.\begin{array}{@{}r@{}r@{}r@{}}\\{}\\{}\\{}\end{array}\color{red}\right\}%
          \color{red}\begin{tabular}{l}We learn\\ the learner \end{tabular}$}}
	            \State {Learn new $(\theta)$}
	            \State {Update the $(F(\theta))$}
	         %\EndWhile
	         \EndFor

	        \State{$\theta'$=$\theta'$+$\nabla \mathscr{L(F(\theta))}$}
	        \rlap{\smash{$\left.\begin{array}{@{}r@{}r@{}r@{}}\\{}\\{}\end{array}\color{red}\right\}%
          \color{red}\begin{tabular}{l}We optimize the meta-learner\\until $\nabla \mathscr{L(F(\theta))} \leadsto 0$.\end{tabular}$}}
	        \State {Update $\mathscr{F(\theta')} \backsim \theta'$ } 
	        \EndWhile
	       \State{\bf{End}}		
	        		
	\end{algorithmic}
	\label{tab: MTL}
\end{algorithm} 
 \subsection{Preliminaries for Zero Shot Learning}
 %here please explain the notation, the background that is needed to understand the algorithm 
 
 \textbf{Notations} Let's suppose  $(\mathscr{D})$ stands for a training dataset  $(\mathscr{D})$=( $\mathscr{X}$, $\mathscr{Y}$) with seen classes $E$=$\{1,2, \cdots , n\}$. Consider $\mathscr{X}$ involving $\mathscr{d}$ dimensions of semantic space with required data, we map $\mathscr{X}$ into k-dimensional latent space with a mapping matrix $\mathscr{W}$. We name this latent representation, ${S}$. Then, we map this latent representation back to feature space $\mathscr{\hat{X}}$ using the transpose of $\mathscr{W}$, which is $\mathscr{W^T}$.
 In line 6 of algorithm \ref{algorithm: MTL-Meta-Mapping}, the authors used a well-known Sylvester equation to calculate an optimum mapping matrix using $\mathscr{A}$, $\mathscr{B}$ and $\mathscr{C}$ which stands for $SS^T$, $\lambda XX^T$ and $(1+\lambda)SX^T$, respectively. We use this $\mathscr{W}$ to recognise unseen classes in a testing dataset, $D^t$=( $X^t$, $Y^t$), with unseen classes $Z$=$\{n+1,n+2, \cdots , m\}$.

\section{Experimental Result}
We investigate extensively to understand the benefits of
a few factors in SAE algorithm \cite{kodirov2017semantic}. SAE only has one parameter called $\lambda$ which is set differently for separate datasets. However, SAE has a few embedded parameters including \textbf{HITK}, Dist and sorting mode, which are required to be optimally tuned. We tune the embedded parameters with different ranges and values as follows: \textbf{HITK} ranges between 1 and the total number of unseen classes per dataset, Dist includes a kernel algorithm to calculate similarity between mapped unseen class instances and learned seen class instances, sorting mode occurs in either ascending or descending order. It is worth mentioning that we mostly give \textbf{HITK} the values of 1$\cdots$7 and 10.
Examining
all possible combinations is quite expensive, therefore, we present a set of results per different value of \textbf{HITK} for comparison. We find out that only  \textbf{HITK}, one of embedded parameters, has direct sensible, publishable and positive effect on result of tuned-SAE performance.
\subsection{Semantic Space}
To accurately calculate performance of tuned-SAE, we describe the semantic space (SS) and dimension of semantic space (SS-D). We directly emphasise on semantic space since the idea behind zero-shot learning relies on this. The way we describe the inputs becomes important in the training phase to ZSL, especially SAE. In previous research studies, scientists applied two different types of semantic space in their work. One is attribute ($A$), one is word vector ($W$). All the work we present in tables \ref{table:TunedSAE-small} and \ref{table:Tuned-SAE-largedataset} mostly used attribute (A), except two research studies.  SJE \cite{akata2015evaluation} worked with a combined semantic space (A+W) that yielded a result, 76.3, which was better than the basic attribute-based method (DAP). Moreover,  SS-Voc\cite{fu2016semi} leveraged both (A/W) but not at the same time, and the performed results, 78.3/68.9, illustrate that their approach had been well-computed.  However, SAE \cite{kodirov2017semantic} only used attribute-based semantic space to calculate the performance of recognising unseen classess. It is noteworthy to say that, only word-vector('W') as SS has been used for large datasets in the compared work illustrated in table \ref{table:Tuned-SAE-largedataset} \cite{kodirov2017semantic}.
\subsection{Pre-defined Parameters}
%..... explain the results of our main reference
In \cite{kodirov2017semantic}, Kodirov \emph{et al} compared their proposed method, SAE, with more than 10 highly qualified methods using small datasets, which include AwA \cite{lampert2013attribute}, CUB \cite{wah2011caltech}, $aP\&Y$ \cite{farhadi2009describing} and SUN \cite{patterson2014sun}. The authors improved the accuracy of recognising unseen images at least $\%6$ in comparison with $SS-voc$ and at most  $\%20$ in comparison with basic attibute-based learning method, DAP.   Further, the researchers used two large datasets: $ImNet-1$ \cite{fu2016semi} and $ImNet-2$ \cite{fu2016semi}. Their image recognition errors for large datasets are beyond $\%60$.
\subsection{Effect of Parameter Tuning on Accuracy of Meta-learning for Computer Vision}
To illustrate in detail the effect of parameter tuning on accuracy of meta-learning for computer vision, first, we discuss the used datasets, provide an ablation study, define an evaluation metric and present the state-of-the-art works and provide comparative evaluation in the following sections.
\subsection{Dataset}
We choose two small, but popular, and one large  benchmark dataset for ZSL  in this study: AwA (Animals with Attributes) \cite{lampert2013attribute} consists
of more than 30,000 images with respect to 50 different classes of animals; CUB-200-2011 Birds (CUB) \cite{wah2011caltech} consists of 218 instances, 1000 seen classes and 360 unseen classes; ImNet-2 \cite{fu2016semi} provides 1000 classes for seen classes and 360 classes for unseen classes, where seen and unseen classes are extracted from ILSVRC2012,  ILSVRC2010, respectively.  Table \ref{tab:dataset} illustrates details information of these datasets either training and testing.
%%%%%%%%%%%%%%%%%%%%%%%%%%%%%%%%%%%%%%%%%%%%%%%%%%%%%%%%%%%%

\begin{table}[t]
\center
  \caption{Comparing the related methods with our contribution for small datasets}
  \vspace{0.5cm}
  \renewcommand{\arraystretch} {1}
  \hspace{-1cm}%\begin{tabular}{|c|c|c|C|C|}
    
    \begin{tabular}{*5c}
    \toprule
    \backslashbox{\textbf{methods}}{\textbf{Datasets}}  %&\multicolumn{1}{c}{SS}
    &\multicolumn{1}{c}{AwA} %& \multicolumn{1}{c}{SUN}
    & \multicolumn{1}{c} {CUB} \\ 
         \hline
      DAP \cite{lampert2013attribute} %&A
      & 60.1& -\\%72.0
      %&-\\
    
     ESZSL \cite{romera2015embarrassingly} &75.3&%82.1
     48.7\\
    SSE \cite{zhang2015zero}&76.3&%82.5
    30.4\\
     JLSE \cite{zhang2016zero} &80.5&41.8\\
     SJE \cite{akata2015evaluation}(A+W) &76.3&%56.1
     50.1\\
     SynC \cite{changpinyo2016synthesized} &72.9&54.4\\
     MLZSC \cite{bucher2016improving} &72.9&54.4\\
     PRZSL\cite{shigeto2015ridge}&80.4&52.4\\
     
     f-VAEGAN-D2 (IND) \cite{Xian_2019_CVPR}&71.1&61.0\\
     f-VAEGAN-D2 (TRAN)\cite{Xian_2019_CVPR}&\textbf{89.8}&\textbf{71.1}\\
     SS-Voc\cite{fu2016semi}(A/W)&78.3/68.9 & -\\
     SAE (W) \cite{kodirov2017semantic} & 84.7&%91.0
     61.4\\
     $SAE (W^T)$ \cite{kodirov2017semantic} & 84.0&%91.5&
     60.9\\
     
      \hline
      Our contribution   &&\\  \hline
      $SAE(W)-1$&  84.7& 61.4\\
      $SAE(W^T)-1$& 84.0& 60.9\\\hline
      $SAE(W)-2$& 74.6 &\textbf{78.0}\\
      $SAE(W^T)-2$& \textbf{88.9}&\textbf{97.4}\\  \hline
      $SAE(W)-3$& \textbf{83.7}&\textbf{85.13}\\
      $SAE(W^T)-3$&\textbf{ 94.4}&\textbf{97.4}\\  \hline
      $SAE(W)-4$& \textbf{91.0}&\textbf{89.9}\\
      $SAE(W^T)-4$&\textbf{ 97.5}&\textbf{97.4}\\  \hline
     $ SAE(W)-5$& \textbf{96.4}&\textbf{92.4}\\
     $ SAE(W^T)-5$& \textbf{99.1}&\textbf{97.4}\\  \hline
     $ SAE(W)-6$& \textbf{99.7}&\textbf{94.1}\\
      $SAE(W^T)-6$&\textbf{ 99.7}&\textbf{97.4}\\  \hline
      $SAE(W)-7$& \textbf{99.5}&\textbf{95.3}\\
      $SAE(W^T)-7$&\textbf{ 99.8 }&\textbf{97.4}\\  \hline
      $SAE(W)-10$&\textbf{100 }&\textbf{ 97.4}\\
      $SAE(W^T)-10$&\textbf{100 }& \textbf{97.4}\\
                    \hline
    \end{tabular}
       \label{table:TunedSAE-small}
  \end{table}
  
\begin{table} [t]
\center
  \caption{Comparing the related methods with our contribution for a large dataset}
  \vspace{0.5cm}
  \renewcommand{\arraystretch} {1}
  \hspace{-1cm}%\begin{tabular}{|c|c|c|C|C|}
    
    \begin{tabular}{*3c}
    \toprule
    \backslashbox{\textbf{methods}}{\textbf{Datasets}}   &\multicolumn{1}{c}{ImNet-2}  &\multicolumn{1}{c}{ImNet-2}\\
         \hline
   
     ConSE\cite{norouzi2013zero}&15.5&15.5\\
     SS-Voc\cite{fu2016semi}&16.8&16.8\\
     
      $SAE(W)$ \cite{kodirov2017semantic} &26.3&26.3\\
      $SAE(W^T)$ \cite{kodirov2017semantic} &27.2 &27.2\\
     
      \hline
      Our solution\\with parameter tuning   & $\lambda$= 5&$\lambda$= 6\\ 
      \hline
      $SAE(W)-1$&12.2&12.2\\
      $SAE(W^T)-1$& 12.9&13.0\\  \hline
      $SAE(W)-2$&17.6&17.6\\
      $SAE(W^T)-2$& 18.3&18.4\\  \hline
      $SAE(W)-3$& 21.2&21.1\\
      $SAE(W^T)-3$& 22.1&22.1\\  \hline
      $SAE(W)-4$&24.0&23.9\\
      $SAE(W^T)-4$&24.9 &24.9\\  \hline
      $SAE(W)-5$& 26.3& 26.3\\
      $SAE(W^T)-5$& 27.2&\textbf{27.3}\\  \hline
      $SAE(W)-6$&\textbf{28.3} &\textbf{28.3}\\
      $SAE(W^T)-6$&\textbf{29.2 }&\textbf{29.3}\\  \hline
      $SAE(W)-7$&\textbf{30.1} &\textbf{30.1}\\
      $SAE(W^T)-7$&\textbf{31.1} &\textbf{31.2}\\  \hline
      $SAE(W)-10$&\textbf{34.8 }& \textbf{34.7}\\
      $SAE(W^T)-10$&\textbf{35.6} &\textbf{ 35.7}\\
                    \hline
    \end{tabular}
       \label{table:Tuned-SAE-largedataset}
  \end{table}
\subsection{Ablation Study}
In this paper, we work jointly with semantic auto-encoder (SAE), which is an advanced supervised clustering algorithm for the specific purpose of zero-shot learning. The main strength of this paper is tuning SAE, which comes from examining the output by updating parameters.  ZSL usually uses a complex projection function, but SAE leverages a linear projection, according to algorithm \ref{algorithm: MTL-Meta-Mapping}.
\subsection{Evaluation Metric}
We compute performance of the tuned-SAE based on the loss function $\mid \mid Pred - Ground_{T}\mid \mid$, which is also presented in \cite{law2016closed} for a metric learning function and supervised clustering.
\subsection{Competitors}
We compare our method, tuned-SAE, with state-of-the-art method \cite{Xian_2019_CVPR} and other work are compared in  \cite{kodirov2017semantic}. All compared research studies have used zero-shot learning (supervised learning) \cite{zhang2016zero} \cite{lampert2013attribute} and semi-supervised learning\cite{fu2016semi}.

\subsection{Comparative Evaluation} 
We make the following observations according to the results in tables \ref{table:TunedSAE-small} and \ref{table:Tuned-SAE-largedataset}: (1) Our tuned-SAE model
obtained the best results on both small and large datasets. (2) On the small  datasets, the gap between tuned-SAE$'$s results and the
strongest competitors are varied due to different results of SAE.
Note that our tuned model is a linear projection function, while most of the compared models use complex nonlinear projection functions, and some of them use more than one semantic space like SJE \cite{akata2015evaluation} and SS-voc \cite{fu2016semi}. (3) Although tuned-SAE performed well on the large-scale dataset ($ImNet-2$), our model did not improve the performance more than $\%7$
in comparison with SAE, but tuned-SAE yields a promising result in comparison with other methods. (4) The performance of tuned-SAE on the small datasets is far better than the large dataset. (5) Last but not least, by increasing $\textbf{HITK}$ value from 2 to 10, our performance directly increases. 
\section{Discussion}
Tuning plays a main role in all algorithms, especially machine learning algorithms. It provides a comprehensive situation for the  algorithms to learn from a training dataset, so the learned model yields a high performance on a testing dataset. However, machine learning algorithms may not do well for unseen classes, although they obtained optimised parameters. In this paper, we address this problem, study semantic auto-encoder (SAE) and develop tuned-SAE in order to gain a better performance on unsupervised clustering, which is one of the approaches for zero-shot learning. The results in table \ref{table:TunedSAE-small} depict that tuned-SAE leads to a better performance than SAE and other related methods. To compare table \ref{table:TunedSAE-small} with table \ref{table:Tuned-SAE-largedataset}, we find that tuned-SAE performs far better for small datasets than large datasets. This paper proves that tuning is a big advantage in zero-shot learning for image recognition. However, it does not work well for large datsets. Table \ref{table:Tuned-SAE-largedataset} shows that with different $\lambda$ values we have have different results for $SAE(W)$ and  $SAE(W^T)$, such that $SAE(W^T)$ plays as a decoder and maps data from semantic space to feature space to compute performance of the work.
\section{Conclusion}
Having an optimal learning model is key in the world of machine learning. However, learning from unseen classes is a critical issue in traditional machine learning. In this paper, we address this problem and investigate advanced learning processes to enable learning from seen classes to predict unseen classes accurately. We aim to focus on SAE as a semantic auto-encoder, which enables us to find an optimal mapping function between semantic space and feature space, such that it also works for unseen semantic space and classes. In this paper, we tune embedded SAE's parameters in a way that SAE yields better results than the original parameters presented in \cite{kodirov2017semantic}. The new results outperform the original results as well as state-of-the-art algorithms.
\bibliographystyle{IEEEtran}
\bibliography{bib.bib}
\end{document}